\documentclass[sigconf]{acmart}
\makeatletter
\def\@ACM@checkaffil{
    \if@ACM@instpresent\else
    \ClassWarningNoLine{\@classname}{No institution present for an affiliation}%
    \fi
    \if@ACM@citypresent\else
    \ClassWarningNoLine{\@classname}{No city present for an affiliation}%
    \fi
    \if@ACM@countrypresent\else
        \ClassWarningNoLine{\@classname}{No country present for an affiliation}%
    \fi
}
\makeatother
\usepackage{epsfig,endnotes}
\usepackage{pifont,hyperref,color,graphicx,dirtree,multirow}
\usepackage{dblfloatfix}
\usepackage{adjustbox}

\usepackage{enumitem}
\usepackage{setspace}
\usepackage{subcaption}
\setlist[itemize]{leftmargin=3mm}

\usepackage{amsmath}
\usepackage{array}
\usepackage{booktabs}
\usepackage{caption}
\usepackage{siunitx}
\usepackage{hyperref}
\usepackage{graphics}
\usepackage{xcolor}

\DeclareMathAlphabet{\mathcal}{OMS}{cmsy}{m}{n}
\SetMathAlphabet{\mathcal}{bold}{OMS}{cmsy}{b}{n}

\newcommand{\bigO}{\mathcal{O}}


\renewcommand\footnotetextcopyrightpermission[1]{} %

\begin{document}

\date{}

\title{Terracorder: Sense Long and Prosper}

\author{Josh Millar, Sarab Sethi, Hamed Haddadi}
\affiliation{%
  \institution{Imperial College London}
}

\author{Anil Madhavapeddy}
\affiliation{%
  \institution{University of Cambridge}
}

\maketitle
\renewcommand{\shortauthors}{Millar et al}
\pagestyle{plain} %

\thispagestyle{empty}

\subsection*{Abstract}
In-situ sensing devices need to be deployed in remote environments for long periods of time; minimizing their power consumption is vital for maximising both their operational lifetime and coverage. We introduce Terracorder – a versatile multi-sensor device – and showcase its exceptionally low power consumption using an on-device reinforcement learning scheduler. We prototype a unique device setup for biodiversity monitoring and compare its battery life using our scheduler against a number of fixed schedules; the scheduler captures more than 80\% of events at less than $50$\% of the number of activations of the best-performing fixed schedule. We then explore how a collaborative scheduler can maximise the \textit{useful} operation of a network of devices, improving overall network power consumption and robustness.
\vspace{-0.1cm}
\section{Introduction}
The low-power operation of battery-powered or energy-harvesting sensor networks is vital for long-term deployments. This includes networks deployed in urban areas; it is often prohibitive to ensure continuous power availability in a deployment with many sensor nodes in diverse locations~\cite{prohibitive}. We focus on event-driven sensing, maximizing network lifetime alongside its \textit{useful} active time. Traditional learning-based approaches for event-driven scheduling require federated/centralized coordination, ongoing data exchange between networked devices, or one \textit{always-on} device in a proximity group \cite{finally}. We instead explore the potential of an on-device reinforcement learning scheduler for event-driven networks, using low-power \textit{collaboration} between neighboring devices to minimize redundancy. We present a first step towards this in the form of an on-device scheduler for isolated devices, whose implementation provides the basis for a large-scale networked approach.


High-quality biodiversity monitoring relies on rich data from power-hungry sensors (i.e. cameras and microphones), yet devices are unlikely to have a continuous power supply (without sacrificing flexibility of deployment) and need to optimize for power consumption. This domain also presents a unique opportunity; learning the schedules of events of interest (bird vocalizations, feeding activity, etc.) is necessary for optimizing network lifetime but also simultaneously generates large-scale information useful to biologists and conservationists.


The concept of a low-power yet multi-sensor device for biodiversity monitoring is frequently discussed, yet remains unrealized \cite{ammod, cammic}. The ideal device is standalone, while also able to collaborate with nearby devices, providing robust and maintainable coverage without redundancy~\cite{manoomin}.
\vspace{-0.1cm}
\subsection{Low-power Hardware} 
Terracorder's advantage lies in its low base power usage, while still being capable of interfacing with multiple power-hungry sensors. The prototype builds on PowerFeather
, an ESP32s3-based module 
featuring ultra-low deep-sleep consumption of $\sim$19$\mu$A (with 3.7V supply). This is significantly lower than other development boards of similar capacity, including other ESP32 variants, and an order of magnitude lower than the Raspberry Pi, on which a vast number of multi-sensor devices are built. Terracorder includes built-in power management features, such as a LiPo/Li-Ion battery charger IC
and fuel gauge,
supporting battery health monitoring and time-to-empty/full battery estimation along with a DC input to source reliable renewable power.
The ESP32s3 itself is based on a XTensa LX7 SoC, with 2MB of quad-SPI PSRAM, 512KB of internal SRAM, and 16KB RTC SRAM for retaining state over deep-sleep. The SoC also includes a RISC-V/FSM ultra-low-power coprocessor for interfacing with external I2C sensors while the main processor remains in deep-sleep. Terracorder provides access to 23 of the ESP32s3’s GPIO for interfacing with external modules, of which 12 are RTC-capable and can be used in deep-sleep for low-power I/O. 
\vspace{-0.1cm}
\subsection{Adaptive Scheduling} 
Terracorder uses adaptive scheduling to manage its low-power modes; user-defined schedules are sufficient for regular event patterns, but lack adaptivity once deployed. There are numerous algorithmic approaches for adaptive scheduling, including exponential backoff, context rule-based and genetic algorithms, and random sampling-based methods that estimate a probability distribution of event occurrences.  However, learning-based schedulers allow for versatile scheduling in the absence of an event model, and become particularly important if deploying many sensors for long durations, as event patterns vary spatiotemporally and performance can degrade if scheduling isn't regularly fine-tuned. We present an on-device reinforcement learning scheduler that optimizes sensor activation rate via cost-balancing of positive and negative activations. This aims at maximizing event detections at minimized operation cost, or under a power constraint, and is the first step towards a collaborative scheduling approach for networked sensors.

Q-learning is a model-free reinforcement learning algorithm that builds a Q-table $Q(s,a)$ of discrete state-action values, and is useful for learning optimal coverage-maximising schedules on energy harvesting devices~\cite{RL1, RL2, smarton}. Q-values are the estimated total discounted reward from taking action $a$ in state $s$ and then following the optimal, learned policy $\pi$. The discounted reward signal is used to update the Q-table:
\vspace{-0.1em}
\begin{equation}
    Q_{\pi}(s,a) = r_0 + \gamma max_{a}Q_{\pi}(\Acute{s}, a)
    \label{eq:1}
\end{equation}

$\gamma$ is a discount factor that balances the importance of immediate versus future reward. Q-learning inference follows a greedy or $\epsilon$-greedy policy, which for a given state picks the maximum Q-value action with probability 1-$\epsilon$ and a random action otherwise. Given device resources are limited, and the scheduler should incur minimal power overhead, Q-learning’s low memory requirements and $\bigO(n)$ inference and update are fitting. These properties are especially important for multi-sensor devices, where different schedules can by used for each component depending on its events of interest. 

We further minimize overhead by \textit{quantizing} the number of discrete actions/states. We divide state into $t$ time periods over which activation rate is fixed. This can be done based on observed/expected event patterns and imposed operational constraints; hourly periods should provide enough granularity for most applications. We also use $K$ possible activation frequencies based on the observed/expected range of event intervals and durations. With hourly periods and $K = 7$ possible actions, the resulting 24x7 Q-table is only $\sim$3KB; multiple Q-tables can therefore be used on one device, for the scheduling of different connected sensors.  

We can formulate the reward $R$ for period $t$ as:
\vspace{-0.3em}
\begin{equation}
R_t = N_{p_t} - w_1 \cdot N_{n_t}
\label{eq:reward}
\end{equation}

$N_p$ is the total number of \textit{positive} activations over $t$ (i.e., those on which an event(s) is detected), and $N_n$ the total \textit{negative} activations. $w_1$ is a tunable parameter for adjusting the scale discrepancy between $N_p$ and $N_n$ to vary event detection priorities (i.e., larger $w_1$ applies more weighting towards minimizing total activations); it can be set based on the range of activation frequencies, and observed or expected event patterns. $w_1$ can also be period-specific, increasing as the number of expected events increases between periods. The reward is designed for broad applicability in various use-cases; it can support continuous periodic recording of fixed duration with an updated action space, more nuanced optimization by adding conditions that influence event occurrence, or apply differentiated rewards for events/actions in different periods. 

The Q-table can be learned based on pre-deployment data and fine-tuned on the device, or learned fully on the device. The latter can initially follow a conservative schedule, or a schedule based on expected event patterns, using occasional random activations to compile a probability distribution of event occurrences for each period $t$. This distribution can be used for Q-table initialization, assigning larger values to state-action pairs corresponding to periods with greater event probabilities, reducing the impact of initially poor scheduling if learning is slow. The Q-table can be updated periodically once converged, or if detection rates fall.
\vspace{-0.5em}
\section{Prototype and Measurements}
We next prototype a Terracorder setup for biodiversity monitoring and validate its long-term sensing capabilities. This application is incredibly important for building an understanding of ecological dynamics; a learning-based scheduler not only enables adaptive scheduling, but also provides useful information for conservationists on large-scale event patterns.  

\begin{figure}[htbp!]
\vspace{-0.2cm}
\begin{center}
\includegraphics[width=8cm]{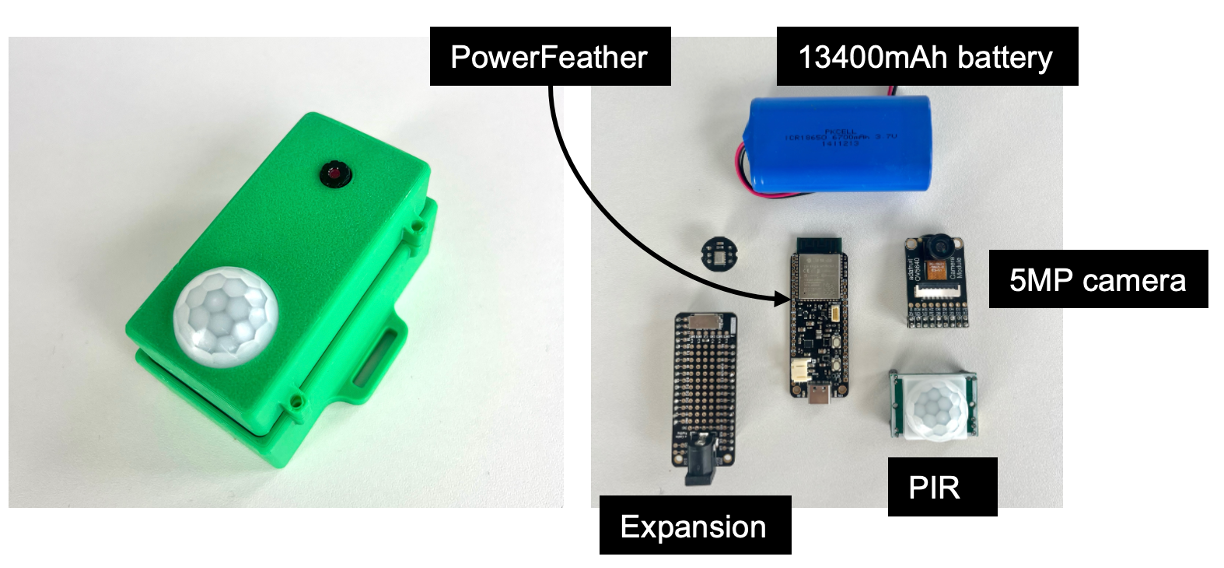}
\vspace{-0.3cm}
\caption{\textbf{Prototype configuration}}\label{fig:prototype}
\end{center}
\vspace{-0.5cm}
\end{figure}

The prototype is configured with a 5MP camera
, omnidirectional I2S microphone
, and PIR sensor
(all shown in Fig. \ref{fig:prototype}). The components are selected for their low-power operation and/or use in existing biodiversity monitoring devices~\cite{pict}. The PIR sensor remains on continuously, including in deep-sleep, acting as an event-trigger for our camera. The microphone is our scheduled component, but recording could also be event-triggered for less frequent events of longer durations.
We measure the current draw of the sensors and device using a high-voltage power monitor capable of $\mu$A-scale measurements.
The board is supplied at 3.3V via its JST-PH battery connector. WiFi is used for the communication of results; this is appropriate for a built environment, however the device can also use 4G or LoRaWAN in non-urban areas.

We perform Q-learning inference and Q-table updates on-device to show their negligible overhead. Goertzel filtering is also performed on-device, for event detection (discussed in \S\ref{section:results}); its measurement follows a 0.1s recording. We also implement an alternative integer-quantized 
one-layer CNN model for event detection, using ESP-TFLite-Micro
; its measurement also follows a 0.1s recording and includes Mel spectrogram generation (also discussed in \S\ref{section:results}).

We attain a low deep-sleep current of <100$\mu$A, with all sensors attached and the RTC enabled (providing deep-sleep timekeeping for scheduling purposes). The power consumption for Q-learning inference and update is $\sim$30$\mu$A and $\sim$70$\mu$A respectively, with a latency of <0.1s, confirming its negligible overhead. Goertzel filtering and TFLite inference (and 0.1s recording) have similar consumptions of $\sim$33.3mA and $\sim$33.1mA respectively, which, given latencies of $\sim$0.03s and $\sim$0.07s, add minimal overhead. However, due to its memory costs, only one quantized TFLite model can realistically be used on-device; Goertzel and other thresholding approaches offer greater scalability. WiFi is the largest power consumer, particularly for the transmission of images, requiring further optimization.

The bare-board active consumption of a Raspberry Pi, as a comparison, can range from $\sim$150mA (Zero) to >500mA (Model 3); a Model 3 setup for acoustic monitoring uses $\sim$300mA on average \cite{sarab}.  PICT, a Pi Zero event-triggered camera setup for biodiversity monitoring, also using a 5MP camera, uses $\sim$150mA in idle operation, and $\sim$374mA with camera/Wifi enabled \cite{pict}. While the sensors used are not entirely equivalent, their power usage highlights an order of magnitude difference between existing devices and ours.

\begin{figure}[h!]
\vspace{-0.2cm}
\begin{center}
    \captionof{table}{PowerFeather current measurements (3.35V in)}
    \vspace{-0.3cm}
    \resizebox{\columnwidth}{!}{%
    \begin{tabular}{l r}
        \toprule
        \textbf{Mode/Operation} & \textbf{Current Draw (mA)} \\ 
        \midrule
        Deep-sleep (PIR and RTC active) & 0.097 \\
        Microphone & \\
        \quad - 3s recording & 31.57\\
        \quad - 0.1s recording + Goertzel filter & 32.34 \\
        \quad - 0.1s recording + TFLite inference & 33.11 \\
        Camera (one activation) & 49.33 \\
        Transmission (via WiFi) & \\ 
        \quad - 3s audio recording & 61.33 \\ 
        \quad - 5MP image & 97.73  \\ 
        QL inference & 0.031 \\ 
        QL update & 0.071 \\ 
        \bottomrule
    \end{tabular}
    }
    \label{tab:energy}
\end{center}
\end{figure}

\begin{figure}[h!]
\vspace{-0.3cm}
\begin{center}
\captionof{table}{Q-learning hyperparameters}
\vspace{-0.2cm}
    \begin{tabular}{lr}
        \toprule
         Parameter & Value \\ \midrule
            $\gamma$ (discount factor) & 0.9 \\
            $\epsilon_{max}$ & 0.3 \\
            $\epsilon_{min}$ & 0.1 \\
            $\epsilon_{decay}$ & 0.99 \\
            $\alpha$ (learning rate) & 0.1 \\
            $\beta$ & $1e^{-5}$ \\
            Actions (seconds) & 3, 5, 60, 300, 1800 \\
        \bottomrule
    \end{tabular}
    \label{tab:parameters}
\end{center}
\vspace{-0.4cm}
\end{figure}
\vspace{-0.1cm}
\section{Isolated Scheduling}
\label{section:results}

We apply an off-the-shelf bird vocalization detection model, BirdNET \cite{birdnet}, to generate event times and durations from the continuous recordings of an acoustic monitoring device in the Stability for Altered Forest Ecosystems (SAFE) project, a large-scale experiment in the Sabah rainforest, Borneo. We use the SAFE data as it is pre-screened yet also temporally heterogeneous in event density.

The generated events are used in building and evaluating our scheduling approach; combined with our measurements from Table \ref{tab:energy}, they provide the basis for estimating the device's battery life in a real-world deployment. We correct for BirdNET false positives by thresholding at a confidence level of 0.7; users can generate species-specific thresholds based on expert validation of pilot data recorded from the sensed environment to further improve detection accuracy \cite{species_specific}.

The schedule is learned over 24 hours of detections and evaluated over the following 24 hours, to imitate day-by-day learning with minimal pre-deployment data. We implement a Goertzel filter on-device to simulate real-time event detection, as commonly used in existing passive acoustic monitoring devices \cite{audiomoth}. This evaluates specific frequency components of a fast Fourier transform (FFT) on buffered recordings, with reduced operations and lower memory usage compared to FFT. We set the bandwidth of multiple filters to cover a range of frequencies relevant to the event(s) of interest, and threshold its median to identify if an audio event is \textit{of interest} and justifies further recording. The Goertzel filter processes a 0.1s recording (at 16000kHz) with just 0.03s latency, minimally impacting our device's lifetime. 

The issue with Goertzel filtering, however, and other related approaches, is its potential for false positives; any event within its range of covered frequencies may be detected (likewise, any marginally off-frequency events would be missed). This can obscure or distort the structure of temporal patterns, resulting in degraded fine-tuning performance. We therefore also implement a single-layer CNN on-device for event detection, trained on Mel spectrograms of 0.1s \textit{positive} slices from BirdNET detections, and \textit{negative} slices from elsewhere in the continuous recordings. This involves extracting Mel spectrograms on-device from buffered recordings using ESP-DSP. 
We attain $\sim$70\% accuracy on unseen event data; however, no validation is done on the positive slices. 

We assume a fixed 0.1s recording period on activation to detect ongoing events; if an event is detected, recording continues for the entirety of the event. Otherwise, the device returns to deep-sleep until its next scheduled activation. The simulation also includes an 0.03s delay based on measured Goertzel filtering latency. We assume that all events have equal priority, although this may not be the case in every practical scenario; to address variations, users can extend our reward to quantify event importance in each period.

We discretize state into hourly periods. Table \ref{tab:parameters} outlines our action space and Q-learning hyperparameters used. The configuration results in a 24x7 Q-table of size $\sim$3KB on the device. 

\begin{figure}[htbp!]
\vspace{-0.3cm}
\begin{center}
\includegraphics[width=7cm]{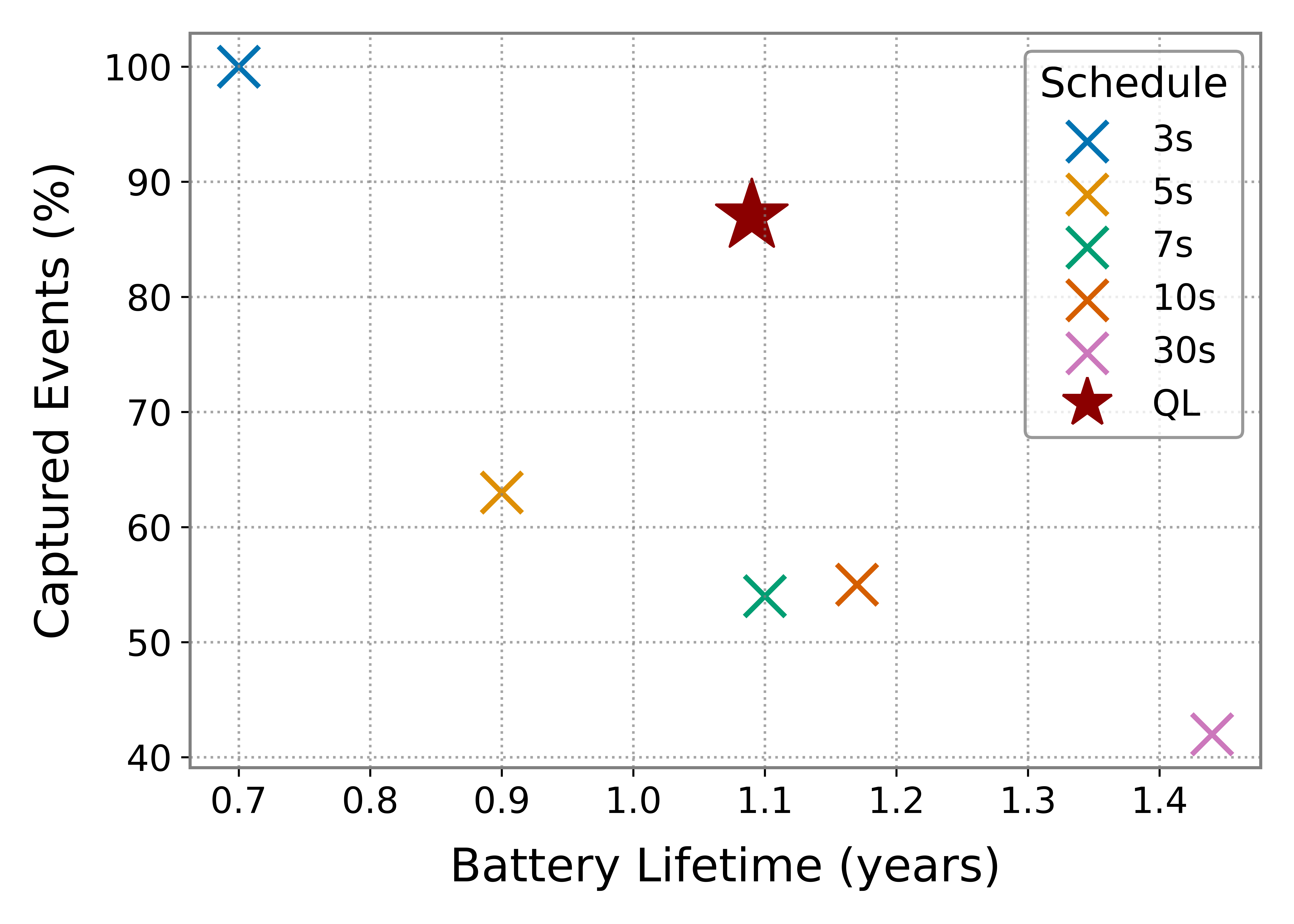}
\vspace{-0.5cm}
\caption{\textbf{Performance of fixed vs. Q-learning schedules.}}\label{fig:2}
\end{center}
\vspace{-0.3cm}
\end{figure}

The scheduling results are visualised in Fig. \ref{fig:2}. We compare our scheduler to a number of fixed baselines that activate the device every $n$ seconds, irrespective of the current period. These baselines are commonly used in real-world biodiversity monitoring deployments, alongside continuous periodic recording of fixed duration  \cite{audiomoth}. However, more complex algorithmic schedules, or ones based on known event patterns, could also be implemented; although these are non-adaptable, they can be used to initialize the Q-table in order to accelerate its convergence. Given the reduced action/state spaces, the Q-learning algorithm can converge in $\sim$30 episodes; this is practical for on-device learning.


The device's lifetime is calculated using a 13400mAh LiPo/Li-Ion battery. We model the average operating current based on our hardware measurements and numbers of events detected. We assume each detected event triggers a 3s recording -- this is based on the average event duration -- and a subsequent Wifi transmission. We also assume one in three events triggers a camera activation and transmission; this is much higher than typical rates for bird/mammal detection in the Sabah rainforest \cite{sabah}. The assumptions made provide an overall relatively conservative estimate of battery lifetime as we {\em (i)} do not parallelize operations (i.e., we use only one core of our dual-core ESP32); {\em (ii)} record the entire duration of each detected event; and {\em (iii)} trigger a camera activation for a large number of detected events. 

Our scheduling approach, using Goertzel filtering, detects 85.3\% of events, increasing battery lifetime by 55.1\% over a fixed 3s baseline that detects all events. This extends lifetime from 0.69 to 1.07 years. The use of TFLite event detection results in similar numbers; 51.9\% improvement at 1.05 years. With a renewable power source such as a solar panel for battery recharging, our lifetime bottleneck shifts from the device's power consumption to its robustness.

\vspace{-0.4em}
\section{Collaborative Scheduling}
\vspace{-0.3em}
\label{section:discussion}

The results outlined are promising, and our scheduler is useful as is for individual deployments or scenarios with constrained inter-device communication. However, most applications, particularly biodiversity monitoring, require large-scale deployments of networked sensors. Given an adequately dense network, the sensing ranges of multiple devices can overlap, and therefore capture the same events. This provides an opportunity for further optimization via collaborative network-wide scheduling; devices can schedule their activations so that they are complementary to each other. This form of network-wide scheduling is often used for maximizing the 
coverage of energy-harvesting sensor networks by minimizing redundant sensing activity. We instead optimise for \textit{useful} active time (i.e., all devices in a sensing radius could be in deep-sleep over periods of low event activity), rather than total active time or coverage. We propose a uniquely collaborative approach, scalable to large networks, that improves the battery-life of each device as more overlapping devices are added.

The scheduling approach presented above needs minimal modification for network-wide scheduling, with its goal now being to maximise the useful active time of the network, rather than its individual devices. This boils down to trying to ensure (1) all events are detected, (2) each event is detected by only one device, and (3) devices are only activated when events occur. 

We modify our reward to the following:
\vspace{-0.2em}
\begin{equation}
R_t = N_{p_t} - w_1 \cdot N_{n_t} - w_2 \cdot \sum^{N}_{i} \left(N_{o_t i} - 1\right) - w_3 \cdot B_{\sigma_t}
\label{eq:2}
\end{equation}
$N_p$ is the number of \textit{positive} activations (i.e. the number of detected events) over all devices for the period $t$, and $N_n$ the number of negative activations. $N_o$ is a list of the number of overlapping activations for each event detected by multiple sensors. $B_{\sigma}$ is the variability of battery levels across the devices; included to discourage the schedule from repeatedly activating the same sensors unless their utility is especially high. $w_1$, $w_2$, and $w_3$ are tunable balancing parameters. Our state space expands to include the activation frequencies alongside the binned number of event detections of each device in period $t-1$. The action-space remains a vector of activation frequencies. Given the increased state/action spaces, network-wide scheduling is best modelled using deep Q-networks, or other deep reinforcement learning approaches. 

However, learning a distributed scheduling policy across all devices leaves on-device coordination impractical; devices operate independently and are unaware of each other's states (e.g. overlapping event detections), and the policy is learned over the entire network-space. This necessitates a federated learning approach to inference, and regular data exchange with a base station or external server, which can be resource-intensive and impractical. 
We instead propose a scheduler that first learns a consistent global policy across the network following eq. \ref{eq:reward} (i.e., ignoring detection overlaps and battery-life variability); inference can be performed locally, at minimal overhead, as it only requires observation of the current period $t$ and its detected events. Then, on initial deployment, groups of mutually overlapping devices establish \textit{clusters} over BLE (or another low-power protocol) using GPS or signal strength. The clusters assign a round-robin style ordering between devices, which rotates every period $t$ \cite{dist_con}. Then, on event detection (or per a number of detections), devices issue a short-range RF ping including a hashed representation of event signal features. This allows each device to estimate the number of overlapping detections in each of its clusters, penalizing their activations appropriately based on eq. \ref{eq:2} ($B_{\sigma}$ is not required), \textit{unless} it's their slot in the round-robin. This approach not only enables on-device localized fine-tuning and minimizes redundant activations of devices in close proximity but also improves the robustness of the network, as devices learn to adapt their schedules if a neighbor breaks down and stops pinging other devices. The approach can also minimize redundant activations in networks following non-adaptive schedules.

We also look to further explore how devices in proximity groups can activate each other via RF (e.g., interpreting durations of RF pulses as specific frequencies), based on event locations and forecasted event patterns. This approach would require inter-device communication to synchronize the number/timing of detections, perform collaborative localization \cite{localization}, and consider battery-life variability, but could further enhance network lifetime and its responsiveness to out-of-schedule events.
\vspace{-0.2cm}
\section{Conclusion} We present Terracorder, a unique biodiversity-focused multi-sensor device with on-device reinforcement learning scheduler. Our approach forms the basis of a collaborative scheduler that significantly improves the useful operation, power consumption, and robustness of networked devices; this then facilitates the large-scale data gathering essential for conservation efforts. 
\vspace{-0.1cm}

{\tiny
\bibliographystyle{acm}
\bibliography{refs}
}

\end{document}